\documentclass[10pt,twocolumn,letterpaper]{article}

\usepackage[pagenumbers]{cvpr} 
\usepackage{bm}

\usepackage{graphicx}
\usepackage{tikz}
\usepackage[accsupp]{axessibility}  
\usepackage{enumitem} 
\usepackage{comment} 
\usepackage{multirow}
\usepackage{pifont}
\usepackage{float}
\usepackage{enumitem}
\setitemize{noitemsep,topsep=0pt,parsep=0pt,partopsep=0pt}
\usepackage{balance}
\usepackage{dsfont}
\usepackage{lipsum}  
\usepackage[thinc]{esdiff}

\usepackage{times}
\usepackage{epsfig}
\usepackage{graphicx}
\usepackage{float}
\usepackage{amsmath}
\usepackage{amssymb}
\usepackage{soul}
\usepackage{multicol}
\usepackage{blindtext}
\usepackage{here}
\usepackage{caption}
\usepackage{makecell}
\usepackage[normalem]{ulem}
\usepackage{multirow}
\usepackage{booktabs}
\usepackage{xcolor, colortbl}
\usepackage{balance}
\usepackage{numprint}
\usepackage{sidecap}

\usepackage{xcolor}
\usepackage{skak}

\renewcommand{\paragraph}[1]{\vspace{0.5em}\noindent\textbf{#1}}

\newcommand{\acronym}{HUGS\xspace}

\newcommand{\smpl}{\mbox{SMPL}\xspace}

\newcommand{\gauss}{3D Gaussians\xspace}

\newcommand{\shapecoeff}{\bm{\beta}}
\newcommand{\shapedim}{{\left| \shapecoeff \right|}}

\newcommand{\shapespaceexpl}{\mathbb{R}^{\shapedim}}

\newcommand{\numjoints}{n_k}

\newcommand{\posecoeff}{\bm{\theta}}
\newcommand{\posedim}{{3\numjoints+3}}

\newcommand{\posespaceexpl}{\mathbb{R}^{\posedim}}

\newcommand{\numverts}{n_v}
\newcommand{\numfaces}{n_t}
\newcommand{\template}{\bar{\bm{T}}}

\definecolor{cvprblue}{rgb}{0.21,0.49,0.74}
\usepackage[pagebackref,breaklinks,colorlinks,citecolor=cvprblue]{hyperref}
\usepackage[capitalize]{cleveref}

\def\paperID{000} 
\def\confName{ArXiv}
\def\confYear{2024}

\begin{document}
\newcommand{\jamie}[1]{\textcolor{red}{#1}}

\title{\acronym: Human Gaussian Splats}

\author{Muhammed Kocabas$^{\symknight\symrook\symbishop}$
\;\; Jen-Hao Rick Chang$^\symknight$ \;\;  James Gabriel$^\symknight$\;\;  Oncel Tuzel$^\symknight$ \;\;  Anurag Ranjan$^\symknight$ \\ \\
$^\symknight$Apple  \;\; $^\symrook$Max Planck Institute for Intelligent Systems \;\; $^\symbishop$ETH Zurich
{}}

\twocolumn[{%
	\renewcommand\twocolumn[1][]{#1}%
	\maketitle
	\begin{center}
		\newcommand{\teaserwidth}{\textwidth}
		\centerline{
			\includegraphics[width=\teaserwidth,clip]{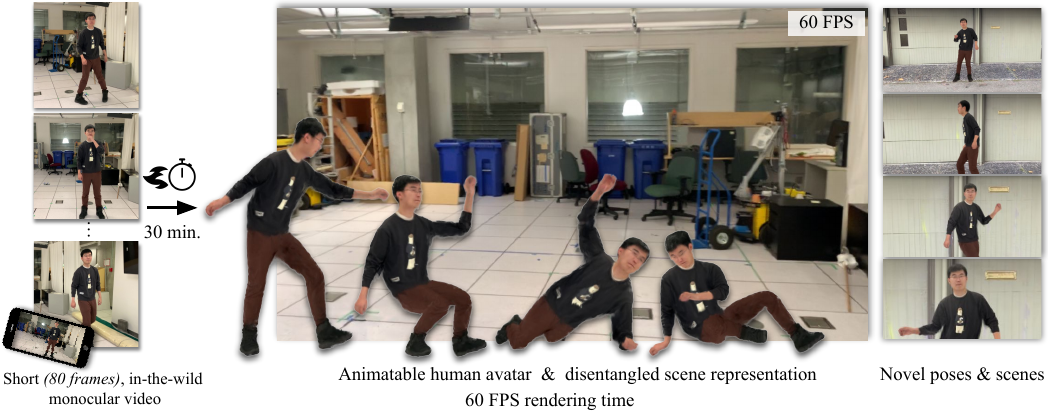}
		}
		\vspace{-1ex}
		\captionof{figure}{\textbf{Human Gaussian Splats (\acronym)} is a neural rendering framework that trains on 50-100 frames of a monocular video containing a human in a scene. HUGS enables novel view rendering with novel human poses at 60 FPS by learning a disentangled representation that can also render the human in other scenes. }
		\label{fig:teaser}
	\end{center}%
}]
\begin{abstract}

Recent advances in neural rendering have improved both training and rendering times by orders of magnitude. While these methods demonstrate state-of-the-art quality and speed, they are designed for photogrammetry of static scenes and do not generalize well to freely moving humans in the environment. In this work, we introduce Human Gaussian Splats (HUGS) that represents an animatable human together with the scene using 3D Gaussian Splatting (3DGS). Our method takes only a monocular video with a small number of (50-100) frames, and it automatically learns to disentangle the static scene and a fully animatable human avatar within 30 minutes. We utilize the \smpl body model to initialize the human Gaussians. To capture details that are not modeled by SMPL (\eg, cloth, hairs), we allow the 3D Gaussians to deviate from the human body model. Utilizing 3D Gaussians for animated humans brings new challenges, including the artifacts created when articulating the Gaussians. We propose to jointly optimize the linear blend skinning weights to coordinate the movements of individual Gaussians during animation. Our approach enables novel-pose synthesis of human and novel view synthesis of both the human and the scene. We achieve state-of-the-art rendering quality with a rendering speed of 60 FPS while being ${\sim}100\times$ faster to train over previous work. Our code will be announced here: \url{https://github.com/apple/ml-hugs}
\end{abstract}

\section{Introduction}

Photorealistic rendering and animation of human bodies is an important area of research with many applications in AR/VR, visual effects, visual try-on, movie production, etc. 
{Early works~\cite{alexander2010emily, alexander2013digitalira, anguelov2005scape} for creating human avatars relied on capturing high-quality data in a multi-camera capture setup, extensive compute, and lots of manual effort.} 
Recent work addresses these problem by directly generating 3D avatars from videos 
using 3D parametric body models like SMPL~\cite{SMPL:2015, pavlakos2019smplx}, which offers advantages such as efficient rasterization and the ability to adapt to unseen deformations. However, the fixed topological structure of parameteric models limit the modeling of clothing, intricate hairstyles and other details of the geometry.

Recent advancements have explored the use of neural fields for modeling 3D human avatars~\cite{guo2023vid2avatar, jiang2022neuman, weng2022humannerf, peng2021neuralbody, bergman2022gnarf, dong2023ag3d}, often using a parametric body models as a scaffold for modeling deformations. Neural fields excel in capturing details like clothing, accessories, and hair, surpassing the quality that can be achieved by rasterization of parametric models with texture and other properties. However, they come with trade-offs, notably being less efficient to train and render.
Furthermore, deformation of neural fields in a versatile manner presents challenges, often requiring recourse to an inefficient root-finding loop, which adversely affects both training and rendering durations~\cite{chen2021snarf,jiang2022instantavatar,yu2023monohuman}.



%

To address these challenges, we introduce a novel avatar representation \emph{\acronym---Human Gaussian Splats}. 
\acronym represents both the human and the scene as 3D Gaussians and utilizing the 3D Gaussian Splatting (3DGS)~\cite{kerbl3Dgaussians} for its improved training and rendering speeds as compared to implicit NeRF representations~\cite{jiang2022neuman, liu2021neuralactor}. 
While utilizing the 3D Gaussian representation allows explicit control of human-body deformation, it also creates new problems. Specifically, a realistic animation of human motions requires a coordination of individual Gaussians to retain surface integrity (\ie, without generating holes or pop outs).

To enable human-body deformation,
we introduce a novel deformation model that represents the human body in a canonical space using the 3D Gaussians.   
The deformation model predicts the mean-shifts, rotations, and scale of the 3D Gaussians to fit the subject's body shape (in the canonical pose).  
%
Moreover, the deformation model predicts the Linear Blend Skinning (LBS) weights~\cite{SMPL:2015} that are used to deform the canonical human into the final pose.
We initialize HUGS from the parameteric \smpl body shape model~\cite{SMPL:2015} but allow the Gaussians to deviate, increase, and pruned from the \smpl model. 
This enables \acronym to model the geometry and appearance details (\eg, hair and clothing) beyond the \smpl model.
The learned LBS weights also coordinate the movement of Gaussians during animation. 
%
HUGS is trained on a single monocular video with 50-100 frames and learns a disentangled representation of the human and scene, enabling versatile use of the avatars in different scenes. 

In summary, our main contributions are 
\begin{itemize}[leftmargin=*]
    \item We propose Human Gaussian Splats (HUGS), a neural representation for a human embedded in the scene that enables novel pose synthesis of the human and novel view synthesis of the human and the scene. 
    \item We propose a novel forward deformation module that represents the target human in a canonical space using \gauss and learns to animate them using LBS to unobserved poses.
    \item HUGS enables fast creation and rendering of animatable human avatars from in-the-wild monocular videos with a small number of (50-100) frames, taking 30 minutes to train, improving over baselines ~\cite{jiang2022neuman, guo2023vid2avatar} by ${\sim}100{\times}$, while rendering at {60 frames per second (FPS)} at HD resolution.\footnote{The train/rendering speed is thanks to 3DGS~\cite{kerbl3Dgaussians}, our contribution is enabling it for deformable cases such as humans.}. 
    \item HUGS achieves state-of-the-art reconstruction quality over baselines such as NeuMan~\cite{jiang2022neuman} and Vid2Avatar~\cite{guo2023vid2avatar} on the NeuMan dataset and the ZJU-Mocap dataset.
    %
\end{itemize}
\section{Related Work}

Early works on photorealistic rendering and animation employed traditional computer graphics pipelines which involved large multi-camera setups such as lightstages ~\cite{debevec2012light} to capture the detailed texture and material of the human body. The animation of human bodies involved the rigging of an artist-created template of a human body mesh~\cite{alexander2013digitalira, alexander2010emily}. The introduction of statistical body shape models~\cite{anguelov2005scape, SMPL:2015, pavlakos2019smplx, STAR:ECCV:2020, SUPR} enabled representation of diverse human shape and animation of the human body by a single model. This reduced the manual effort in creating template meshes and rigging them.
However, these shape models do not account for many details such as clothing, hair, accessories etc. Follow up works such as DRAPE~\cite{drape2012guan} or CAPE~\cite{ma2020cape} augment the shape models to add an additional layer of clothing or altogether choose a different representation such as occupancy~\cite{saito2019pifu, saito2020pifuhd, chen2021snarf, xiu2023econ, xiu2022icon} to represent the details of the geometry. This led to improved estimation of geometry, however the capturing appearance without large capture setups still remained a challenge.

In recent years, Neural Radiance Fields (NeRF)~\cite{mildenhall2020nerf} have enabled a joint representation of geometry and appearance for view-synthesis using multiview images without the need of a large capture setup. Although, a NeRF is designed for capturing static objects, recent work~\cite{peng2021neuralbody, weng2020vid2actor, liu2021neuralactor, weng2022humannerf, jiang2022neuman, Su21arxiv_A_NeRF, guo2023vid2avatar, Feng2022scarf,Mihajlovic:KeypointNeRF:ECCV2022} has extended the NeRF to enable capturing a dynamic moving humans
Weng et al.~\cite{weng2022humannerf} propose a method to model a NeRF representation of a human using a single monocular video enabling 360 degree view generation of a human. Furthermore, NeuMan~\cite{jiang2022neuman} introduces a joint NeRF representation of human and the scene capable of view synthesis and animation of the human in the scene. However, a major limitation of NeRF-based methods is that NeRFs are slow to train and render. Several methods have emerged to speed up training and rendering of NeRFs. These include using an explicit representation such as learning a function at grid points \cite{chen2022tensorf, reiser2021kilonerf}, using hash encoding \cite{muller2022instantngp} or altogether discarding the learnable component~\cite{fridovich2022plenoxels,liu2020neural}. 

Recent work on 3D Gaussian Splatting~\cite{kerbl3Dgaussians} uses a set of 3D Gaussians to represent a scene and renders it by splatting and rasterizing the Gaussians. This approach significantly improves the training and rendering times over traditional NeRFs. Recent work has addressed the extension of 3DGS scenes to controlled dynamic scenes~\cite{wu20234dgs} and multi-camera capture setup~\cite{luiten2023dynamicgs}. However, the 3D Gaussian Splatting framework is not trivial to extend to dynamic humans that allows for both novel-view and novel-pose synthesis of human and the scene.

Our methods builds on the 3D Gaussian Splatting framework~\cite{kerbl3Dgaussians} and utilize the \smpl body shape model~\cite{SMPL:2015} as a prior and learns a deformation model for animation control. 
We use a triplane and three MLPs to coordinate the Gaussians (\eg, their rotation, scale, color, and LBS weights). 
%




\section{Method}
\begin{figure*}[t]
    \centering
    \includegraphics[width=\linewidth]{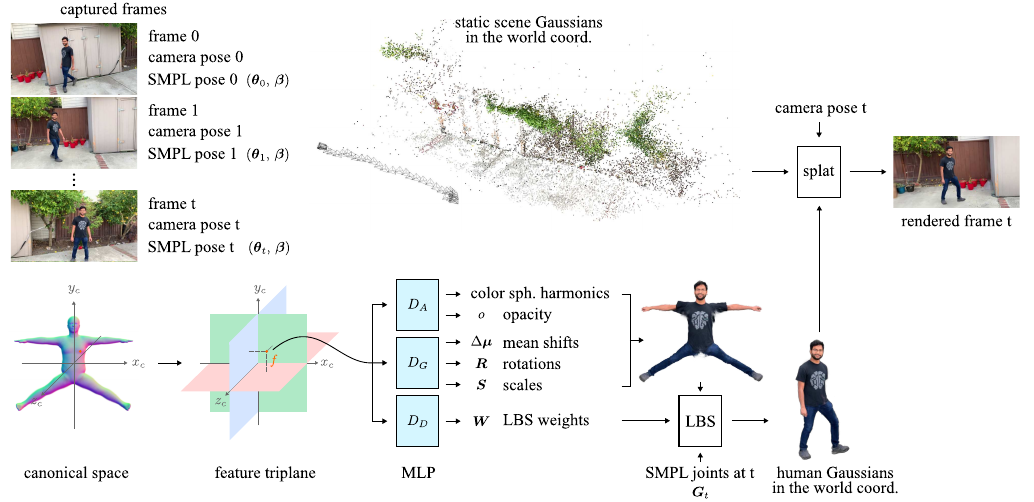}
    \vspace{-7mm}
    \caption{\textbf{\acronym overview.} Given a video with dynamic human and camera motions, \acronym recovers an animatable human avatar and synthesizes human and scene from novel view points. 
    Our method represents both the human and the scene as 3D Gaussians. The human Gaussians are parameterized by their mean locations in a canonical space and the features from a triplane. Three MLPs are used to estimate their color, opacity, additional shift, rotation, scale, and LBS weights to animate the Gaussians with given joint configurations.  
    The human and the scene Gaussians are combined and rendered together with splatting. 
    } 
    \label{fig:overview}
    \vspace{-2ex}
\end{figure*}{}

Given a monocular video containing camera motions, a moving human, and a static scene, our method automatically disentangles and represents the human and the static scene with 3D Gaussians.
The human Gaussians are initialized using the \smpl body model and the scene Gaussians are initialized from the structure-from-motion point cloud from COLMAP~\cite{colmapschoenberger2016mvs, colmapschoenberger2016sfm}.
In the following, we first quickly review 3D Gaussian splatting and the \smpl body model. Then, we introduce the proposed method to address challenges when modeling and animating humans in the 3D Gaussian framework.  

\subsection{Preliminaries}

\paragraph{3D Gaussian Splatting (3DGS)~\cite{kerbl3Dgaussians}}
%
represents a scene by arranging 3D Gaussians. 
The $i$-th Gaussian is defined as 
\begin{equation}
    G(\mathbf{p}) = o_i \, e^{-\frac{1}{2} (\mathbf{p} - \bm\mu_i)^T \bm\Sigma_{i}^{-1} (\mathbf{p} - \bm\mu_i)},
    \label{eq:gauss3d}
\end{equation}
where $\mathbf{p} \, {\in} \, \mathbb{R}^3$ is a xyz location, $o_i \, {\in} \, [0, 1]$ is the opacity modeling the ratio of radiance the Gaussian absorbs, $\bm\mu_i \, {\in} \, \mathbb{R}^3$ is the center/mean of the Gaussian, and the covariance matrix $\bm\Sigma_{i}$ is parameterized by the scale $\mathbf{S}_{i} \, {\in} \, \mathbb{R_+}^3$ along each of the three Gaussian axes and the rotation $\mathbf{R}_{i} \, {\in} \, SO(3)$ with $\bm\Sigma_{i} = \mathbf{R}_{i} \mathbf{S}_{i} \mathbf{S}_{i}^\top \mathbf{R}_{i}^\top$.
Each Gaussian is also paired with spherical harmonics~\cite{ramamoorthi2001efficient} to model the radiance emit towards various directions. 

During rendering, the 3D Gaussians are projected onto the image plane and form 2D Gaussians~\cite{zwicker2001surface} with the covariance matrix $\bm\Sigma_{i}^{\textrm{2D}} = \bm{J} \bm{W} \bm\Sigma_{i} \bm{W}^\top \bm{J}^\top$,
%
%
where $\bm{J}$ is the Jacobian of the affine approximation of the projective transformation and $\bm{W}$ is the viewing transformation. 
%
The color of a pixel is calculated via alpha blending the $N$ Gaussians contributing to a given pixel: 
\begin{equation}
    C = \sum_{j = 1}^{N} c_j \alpha_{j} \prod_{k=1}^{j-1} (1 - \alpha_{k}),
\end{equation}
where the Gaussians are sorted from close to far, $c_j$ is the color obtained by evaluating the spherical harmonics given viewing transform $W$, and $\alpha_j$ is calculated from the 2D Gaussian formulation (with the covariance $\bm\Sigma_{j}^{2D}$) multiplied by its opacity $o_j$.
The rendering process is differentiable, which we take advantage of to learn our human model.


%
%

\paragraph{SMPL~\cite{SMPL:2015}} is a parametric human body model which allows pose and shape control. 
The \smpl model comes with a template human mesh $(\template, \bm{F})$ in the rest pose (\ie, T-pose) in the template coordinate space. 
$\template \, {\in} \, \mathbb{R}^{\numverts \times 3}$ are the $\numverts$ vertices on the mesh, and $\bm{F} \, {\in} \, \mathbb{N}^{\numfaces\times 3}$ are the $\numfaces$ triangles with a fixed topology. 
Given the body shape parameters, $\shapecoeff \, {\in} \, \shapespaceexpl$, and the pose parameters, $\posecoeff \, {\in} \, \posespaceexpl$, \smpl transforms the vertices $\template$ from the template coordinate space to the shaped space via
\begin{equation}
    T_S(\shapecoeff, \posecoeff) = \template + B_{S}(\shapecoeff) +  B_{P}(\posecoeff),
    \label{eq:smpl vertex}
\end{equation}
where $T_S(\shapecoeff, \posecoeff)$ are the vertex locations in the shaped space, $B_{S}(\shapecoeff) \, \in \, \mathbb{R}^{\numverts \times 3}$ and $B_{S}(\posecoeff) \, \in \, \mathbb{R}^{\numverts \times 3}$ are the xyz offsets to individual vertices.
The mesh in the shaped space fits the identity (\eg, body type) of the human shape in the rest pose. 
%
%
To animate the human mesh to a certain pose (\ie, transforming the mesh to the posed space), \smpl utilizes $\numjoints$ predefined joints and Linear Blend Skinning (LBS).
The LBS weights $\bm{W} \, {\in} \, \mathbb{R}^{\numjoints {\times} \numverts}$ are provided by the \smpl model.
Given the $i$-th vertex location on the resting human mesh, $\bm{p}_i \, {\in} \, \mathbb{R}^3$, and individual posed joints' configuration (\ie, their rotation and translation in the world coordinate), $\bm{G} = [\bm{G}_1, \dots, \bm{G}_{\numjoints}]$, where $\bm{G}_k \, {\in} \, SE(3)$, the posed vertex location $\bm{v}_i$ is calculated as $\bm{v}_i = \left( \sum_{k=1}^{n_k} W_{k,i} \, \bm{G}_k \right) \bm{p}_i$, 
%
%
where $W_{k,i} \, {\in} \, \mathbb{R}$ is the element in $\bm{W}$ corresponding to the $k$-th joint and the $i$-th vertex. 
While the \smpl model provides an animatable human body mesh, it does not model hair and clothing. 
Our method utilizes \smpl mesh and LBS only during the initialization phase and allows Gaussians to deviate from the human mesh to model details like hairs and clothing. 
%
%

\subsection{Human Gaussian Splats}

Given $T$ captured images and their camera poses, we first use a pretrained \smpl regressor~\cite{goel2023humans4d} to estimate the \smpl pose parameters for each image, $\bm{\theta}_1, \dots, \bm{\theta}_T$, and the body shape parameters, $\bm\beta$, that is shared across images.\footnote{We also obtain a coordinate transformation from \smpl's posed space to the world coordinate (used by the camera poses) for each frame, following Jiang \etal~\cite{jiang2022neuman}. For simplicity, we will ignore the coordinate transformation in the discussions.} 
Our method represents the human with 3D Gaussians and drive the Gaussians using a learned LBS. 
Our method outputs the Gaussian locations, rotations, scales, spherical harmonics coefficients, and their LBS weights with respect to the $\numjoints$ joints. 
%
%
%
An overview of our method is illustrated in \cref{fig:overview}.

The human Gaussians are constructed from their center locations in a canonical space, a feature triplane~\cite{Peng2020ECCV,Chan2022} $\bm{F} \, {\in} \, \mathbb{R}^{3{\times}h{\times}w{\times}d}$, and three Multi-Layer Perceptrons (MLPs) which predict properties of the Gaussians.
All of them are optimized per person.
The Human Gaussians live in a canonical space, which is a posed space of \smpl where the human mesh performs a predefined Da-pose.

\paragraph{Rendering process.} Given a joint configuration $\bm G$, to render an image, for each Gaussian, we first interpolate the triplane at its center location $\bm{\mu}_i$ and get feature vectors $\bm{f}_x^i, \bm{f}_y^i, \bm{f}_z^i \, {\in} \, \mathbb{R}^d$.
The feature $\bm{f}^i$ representing the $i$-th Gaussians is the concatenation of $\bm{f}_x^i, \bm{f}_y^i, \bm{f}_z^i$.
Taking $\bm{f}^i$ as input, an appearance MLP, $D_A$, outputs the RGB color and the opacity of the $i$-th Gaussian; a geometry MLP, $D_G$, outputs an offset to the center location, $\Delta \bm{\mu}_i$, the rotation matrix $\bm{R}_i$ (parameterized by the first two columns), and the scale of three axes $\bm{S}_i$; a deformation MLP, $D_D$, outputs the LBS weights, $\bm{W}_i \, {\in} \, \mathbb{R}^{\numjoints}$ for this Gaussian.
The LBS uses $\bm{W}$ and the joint transformation $\bm G$ to transform the Human Gaussians, which are then combined with the Scene Gaussians and splat onto the image plane.
The rendering process is end-to-end differentiable.

\paragraph{Optimization.} We optimize the center locations of the Gaussians $\bm{\mu}$, the feature triplane, and the parameters of the three MLPs.\footnote{We also follow Jiang \etal~\cite{jiang2022neuman} and adjust the per image \smpl pose parameters $\bm{\theta}$ during the optimization, since $\bm{\theta}$ are initialized by an off-the-shelf \smpl regressor~\cite{goel2023humans4d} and may contain errors (see the details in the supplementary material).}
The rendered image is compared with the ground-truth captured image using $\mathcal{L}_1$ loss, the SSIM loss~\cite{ssim} $\mathcal{L}_{\text{ssim}}$, and the perceptual loss~\cite{simonyan2015vgg} $\mathcal{L}_{\text{vgg}}$. 
%
We also render a human-only image (using only the human Gaussians on a random solid background) and compare regions containing the human in the ground-truth image using the above losses.
The human regions are obtained using a pretrained segmentation model~\cite{kirillov2023segmentanything}. 
We also regularize the learned LBS weights $\bm{W}$ to be close to those from SMPL with an $\ell_2$ loss.
Specifically, to regularize the LBS weights $\bm{W}$, for individual Gaussians we retrieve their $k=6$ nearest vertices on the \smpl mesh and take a distance-weighted average of their LBS weights to get $\hat{\bm{W}}$. The loss is $\mathcal{L}_{\text{LBS}} = \| \bm{W} - \hat{\bm{W}} \|_{\text{F}}^2$.

%
%
Specifically, our loss is composed of
\begin{multline}
\mathcal{L} = \underbrace{\lambda_1 \mathcal{L}_1 + \lambda_2 \mathcal{L}_{\text{ssim}} + \lambda_3 \mathcal{L}_{\text{vgg}}}_{\text{scene + human}} \\ + \underbrace{\lambda_1 \mathcal{L}^h_1 + \lambda_2 \mathcal{L}^h_{\text{ssim}} + \lambda_3 \mathcal{L}^h_{\text{vgg}}}_{\text{human}} + \lambda_4 \mathcal{L}_{\text{LBS}},
\end{multline}
where $\lambda_1 = 0.8$, $\lambda_2 = 0.2$, $\lambda_3 = 1.0$, $\lambda_4 = 1000$ for all scenes in the experiments.
We employ the Adam optimizer~\cite{KingBa15} with a learning rate of $10^{-3}$, coupled with a cosine learning rate schedule.

We initialize the center of the Gaussians, $\bm\mu$, at the canonical-posed \smpl mesh vertices (so we have the same number of Gaussians as the \smpl vertices at the beginning of the optimization). We pretrain the feature triplane and the MLPs to output RGB color as $[0.5, 0.5, 0.5]$, opacity $o = 0.1$, $\Delta \bm{\mu} = 0$, rotation $\bm{R}$ so that z-axis of the Gaussians align with the corresponding \smpl vertex normal, scale $\bm{S}$ as the average incoming edges' lengths, and LBS weights $\bm{W}$ as those from \smpl (since the Gaussians lie exactly on the \smpl vertices).
The pretraining takes 5000 iterations (1 minute on a  GeForce 3090Ti GPU).
We use an upsampled version of \smpl with $\numverts \, {=} \, 110,210$ vertices and $\numfaces \, {=} \, 220,416$ faces.
Note that the \smpl mesh and LBS weights are only used in the initialization and regularization, \ie, they are not used during testing.

During the optimization, similar to the standard 3DGS, we clone, split, and prune Gaussians, every 600 iterations, based on their loss gradient and opacity.
They are important steps to avoid local minima during the optimization.
To clone and split, we simply add additional entries in $\bm{\mu}$ by repeating existing centers (cloning) and randomly sampling within the Gaussians with respect to their current shapes (\ie, $\bm{R}$ and $\bm{S}$) (splitting).
To prune a Gaussian, we remove it from $\bm{\mu}$.
Since the new Gaussians' centers are close to the original ones on the triplane, their features are similar and thus the new Gaussians have similar shape as the originals, allowing the optimization to proceed normally.
To make the split Gaussians smaller, we record a base scale $s \in \mathbb{R}_+$ for each Gaussian. The base scale is initialized as 1 and every time a Gaussian is split, we divide the base scale by 1.6. The actual scale of the Gaussian is $s$ multiplied with the MLP estimate $\bm{S}$.
The entire optimization takes 12K iterations, and 30 minutes on a GeForce 3090Ti GPU.  At the end of the optimization, the human is represented by 200K Gaussians on average.

\paragraph{Test-time rendering.} Importantly, after the optimization, the 3D Gaussians can be explicitly constructed, allowing direct animation of the human Gaussians using the LBS weights.
In other words, we \textit{do not need to evaluate} the triplane and MLPs to render new human poses. 
This is a big advantage compared to methods that represent human as implicit neural fields.

\begin{figure*}[t]
    \centering
    \includegraphics[width=\linewidth]{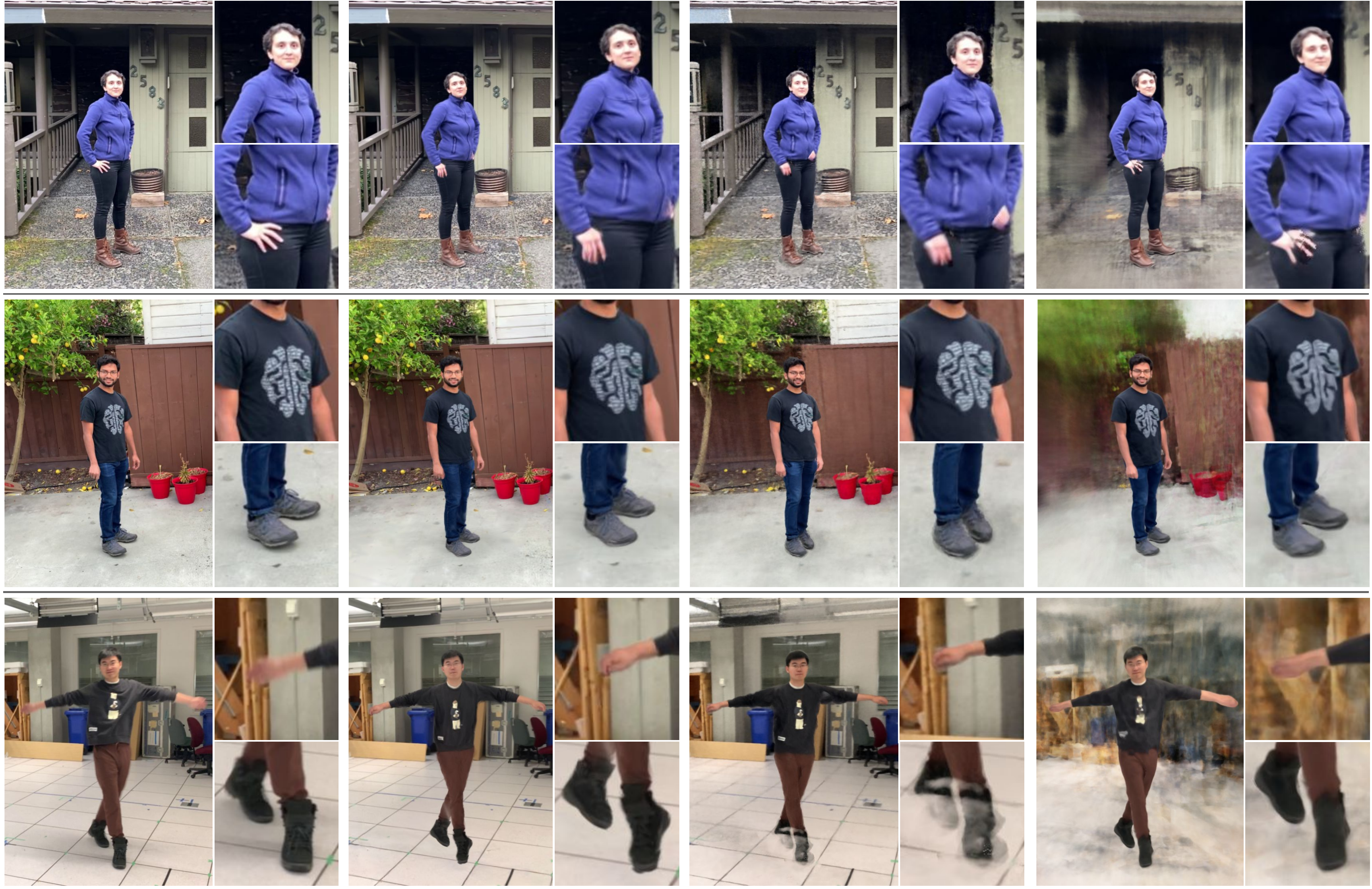}
    \small
    \begin{tabular}{cccc}
         \quad Ground Truth \qquad \qquad & \qquad \qquad HUGS (ours) \qquad \qquad & \qquad \quad \qquad NeuMan~\cite{jiang2022neuman} \qquad \quad & \qquad \quad \qquad Vid2Avatar~\cite{guo2023vid2avatar} \qquad 
    \end{tabular}
    \vspace{-3mm}
    \caption{Qualitative results comparing HUGS (ours) with NeuMan and Vid2Avatar with full human (left) and zoomed-in regions (right) for each of the methods. HUGS shows better reconstruction quality especially around hands, feet and clothing wrinkles.} 
    \label{fig:qualitative_sota}
\end{figure*}{}
\begin{figure*}
    \centering
    \includegraphics[width=\linewidth]{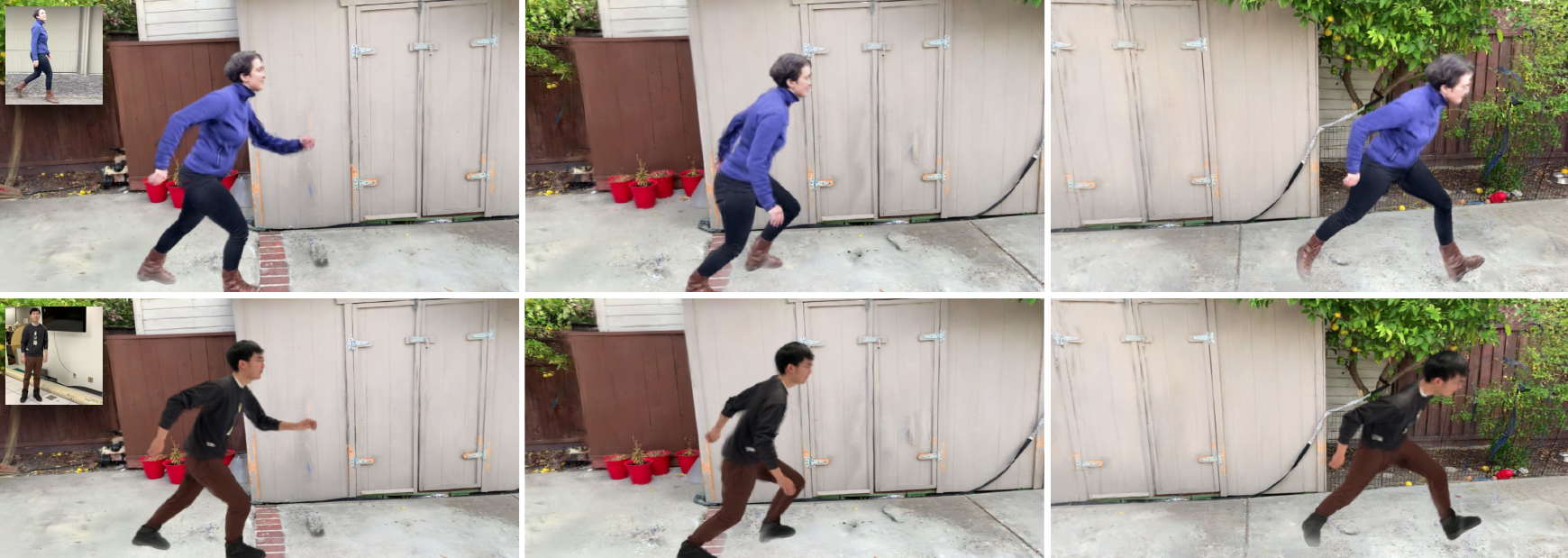}
    \caption{Rendering obtained by transferring the Human Gaussians to a different scene. Top-left corner shows the original scene in which the human was captured.}
    \label{fig:composition}
\end{figure*}{}

\section{Experiments}
\subsection{Datasets}

\paragraph{NeuMan Dataset~\cite{jiang2022neuman}} consists of six videos, each lasting between 10 to 20 seconds, featuring a single individual captured using a mobile phone. The camera pans through the scenes, facilitating multi-view reconstruction. The sequences are denoted as Seattle, Citron, Parking, Bike, Jogging, and Lab. Following the approach outlined in~\cite{jiang2022neuman}, we split frames into 80\% training frames, 10\% validation frames, and 10\% test frames.

\paragraph{ZJU-MoCap Dataset~\cite{peng2021neuralbody}} consists of videos of a human captured in a lab using multi-view capture setup. To align with the methodology in~\cite{weng2022humannerf,yu2023monohuman}, we select six subjects (377, 386, 387, 392, 393, 394) showcasing diverse motions. We employed images captured by "camera 1" as input and utilized the other 22 cameras for evaluation. The camera matrices, body pose, and segmentation provided by the dataset were directly applied in our evaluation process.

\subsection{Qualitative Results}
\label{sec:qualitative}
\paragraph{State-of-the-art Comparison.} We show the qualitative results of our method in ~\cref{fig:qualitative_sota} and compare it with Vid2Avatar~\cite{guo2023vid2avatar} and NeuMan~\cite{jiang2022neuman}. The results are shown from the test samples of the NeuMan dataset~\cite{jiang2022neuman} that are not seen during training.
In the scene background regions, HUGS shows better reconstruction quality than both Vid2Avatar and NeuMan. Vid2Avatar shows blurry scene reconstruction with several artifacts. In contrast, NeuMan shows better scene reconstruction quality but misses fine details such as the house numbers (zoomed-in) in the first row, the wooden plank (zoomed-in) in the second row and the cupboard (zoomed-in) in the third row. In comparison, HUGS shows better reconstruction quality and preserves these fine details as shown in the zoomed-in regions.

In the human regions, Vid2Avatar shows artifacts in the hand region (row 1) and blurry reconstruction in the feet (row 2) and arm region (row 3). In contrast, NeuMan gets better details of the feet regions in some cases (row 2) and introduces artifacts in hands (row 2) and feet (row 3) regions in other cases. In comparison, our method preserves the details around hand and feet and shows better reconstruction quality. Furthermore, our method also preserves the structure around clothing (row 1) where the wrinkles are reconstructed well while preserving the structure of the zipper (zoomed-in) around it compared to previous work.

In summary, we note that HUGS shows better reconstruction quality of both the scene and the human as compared to previous methods while being orders of magnitude faster to train and render (see \S \ref{sec:timing} for speed comparison). We will provide additional qualitative results with videos in the Supp. Mat.

\begin{figure}[t]
    \centering
    \includegraphics[width=\linewidth]{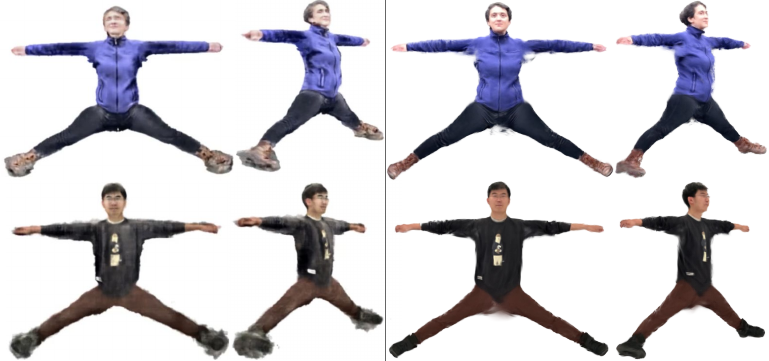}
    \begin{tabular}{cc}
         NeuMan~\cite{jiang2022neuman} \quad \quad & \qquad \qquad HUGS (ours) 
    \end{tabular}
    \caption{Visualization of Human in canonical Da-pose for HUGS (ours) showing qualitative improvements over NeuMan~\cite{jiang2022neuman}.} 
    \label{fig:canonical}
\end{figure}
\paragraph{Canonical Human Shapes.} In ~\cref{fig:canonical}, we show the reconstruction of the human in the canonical space. We note that our method captures fine details around the feet and hands of the human which look noisy in the case of NeuMan~\cite{jiang2022neuman}. Furthermore, we note that our method preserves rich details on the face. This enables us to achieve high reconstruction quality during the animation phase. 

\definecolor{tabfirst}{rgb}{1, 0.7, 0.7}
\definecolor{tabsecond}{rgb}{1, 0.85, 0.7}
\definecolor{tabthird}{rgb}{1, 1, 0.7}

\begin{table*}[]
    \centering
    \resizebox{\textwidth}{!}{
    \begin{tabular}{c|ccc|ccc|ccc|ccc|ccc|ccc}
    \toprule
        & \multicolumn{3}{c|}{\textbf{Seattle}} & \multicolumn{3}{c|}{\textbf{Citron}} & \multicolumn{3}{c|}{\textbf{Parking}} & \multicolumn{3}{c|}{\textbf{Bike}} & \multicolumn{3}{c|}{\textbf{Jogging}} & \multicolumn{3}{c}{\textbf{Lab}}   \\
    \midrule
        & PSNR $\uparrow$ & SSIM $\uparrow$ & LPIPS $\downarrow$ & PSNR $\uparrow$ & SSIM $\uparrow$ & LPIPS $\downarrow$ & PSNR $\uparrow$ & SSIM $\uparrow$ & LPIPS $\downarrow$ & PSNR $\uparrow$ & SSIM $\uparrow$ & LPIPS $\downarrow$ & PSNR $\uparrow$ & SSIM $\uparrow$ & LPIPS $\downarrow$ & PSNR $\uparrow$ & SSIM $\uparrow$ & LPIPS $\downarrow$  \\
    \midrule
    NeRF-T &        21.84       &      0.69              &       0.37           &               12.33  &          0.49            &        0.65             &       21.98            &             0.69        &        0.46    &   21.16     &      0.71         &     0.36         &     20.63        &       0.53         &         0.49        &     20.52         &    0.75      &       0.39                \\
    HyperNeRF &    16.43            &     0.43         &    0.40            &     16.81            &    0.41        &          0.56        &      16.04           &           0.38      &         0.62    &     17.64     &   0.42   &      0.43  &       18.52          &  0.39    & 0.52               &      16.75      &    0.51      &     0.23       \\

    Vid2Avatar &  \cellcolor{tabthird}17.41 &  \cellcolor{tabthird}0.56 &  \cellcolor{tabthird}0.60 &  \cellcolor{tabthird}14.32 &  \cellcolor{tabthird}0.62 &  \cellcolor{tabthird}0.65 &  \cellcolor{tabthird}21.56 &  \cellcolor{tabthird}0.69 &  \cellcolor{tabthird}0.50 &  \cellcolor{tabthird}14.86 &  \cellcolor{tabthird}0.51 &  \cellcolor{tabthird}0.69 &  \cellcolor{tabthird}15.04 &  \cellcolor{tabthird}0.41 &  \cellcolor{tabthird}0.70 &  \cellcolor{tabthird}13.96 &  \cellcolor{tabthird}0.60 &  \cellcolor{tabthird}0.68 \\
    NeuMan     & \cellcolor{tabsecond}23.99 & \cellcolor{tabsecond}0.78 & \cellcolor{tabsecond}0.26 & \cellcolor{tabsecond}24.63 & \cellcolor{tabsecond}0.81 & \cellcolor{tabsecond}0.26 & \cellcolor{tabsecond}25.43 & \cellcolor{tabsecond}0.80 & \cellcolor{tabsecond}0.31 &  \cellcolor{tabfirst}25.55 & \cellcolor{tabsecond}0.83 & \cellcolor{tabsecond}0.23 & \cellcolor{tabsecond}22.70 & \cellcolor{tabsecond}0.68 & \cellcolor{tabsecond}0.32 & \cellcolor{tabsecond}24.96 & \cellcolor{tabsecond}0.86 & \cellcolor{tabsecond}0.21 \\
    \midrule
    HUGS       &  \cellcolor{tabfirst}25.94 &  \cellcolor{tabfirst}0.85 &  \cellcolor{tabfirst}0.13 &  \cellcolor{tabfirst}25.54 &  \cellcolor{tabfirst}0.86 &  \cellcolor{tabfirst}0.15 &  \cellcolor{tabfirst}26.86 &  \cellcolor{tabfirst}0.85 &  \cellcolor{tabfirst}0.22 & \cellcolor{tabsecond}25.46 &  \cellcolor{tabfirst}0.84 &  \cellcolor{tabfirst}0.13 &  \cellcolor{tabfirst}23.75 &  \cellcolor{tabfirst}0.78 &  \cellcolor{tabfirst}0.22 &  \cellcolor{tabfirst}26.00 &  \cellcolor{tabfirst}0.92 &  \cellcolor{tabfirst}0.09
    \\
    \bottomrule
    \end{tabular}  
    }
    \caption{Comparison of HUGS (ours) with previous work on test images of the NeuMan dataset~\cite{jiang2022neuman} using PSNR, SSIM and LPIPS metrics. HUGS achieves state-of-the-art performance across every category with the exception of PSNR on the \textit{Bike} sequence.}
    \label{tab:neuman_human_scene}
\end{table*}
\definecolor{tabfirst}{rgb}{1, 0.7, 0.7}
\definecolor{tabsecond}{rgb}{1, 0.85, 0.7}
\definecolor{tabthird}{rgb}{1, 1, 0.7}

\begin{table*}[htb!]
    \centering
    \resizebox{\textwidth}{!}{
    \begin{tabular}{c|ccc|ccc|ccc|ccc|ccc|ccc}
    \toprule
        & \multicolumn{3}{c|}{\textbf{Seattle}} & \multicolumn{3}{c|}{\textbf{Citron}} & \multicolumn{3}{c|}{\textbf{Parking}} & \multicolumn{3}{c|}{\textbf{Bike}} & \multicolumn{3}{c|}{\textbf{Jogging}} & \multicolumn{3}{c}{\textbf{Lab}}   \\
    \midrule
        & PSNR $\uparrow$ & SSIM $\uparrow$ & LPIPS $\downarrow$ & PSNR $\uparrow$ & SSIM $\uparrow$ & LPIPS $\downarrow$ & PSNR $\uparrow$ & SSIM $\uparrow$ & LPIPS $\downarrow$ & PSNR $\uparrow$ & SSIM $\uparrow$ & LPIPS $\downarrow$ & PSNR $\uparrow$ & SSIM $\uparrow$ & LPIPS $\downarrow$ & PSNR $\uparrow$ & SSIM $\uparrow$ & LPIPS $\downarrow$ \\
    \midrule 
    Vid2Avatar &  \cellcolor{tabthird}16.90 &  \cellcolor{tabthird}0.51 &  \cellcolor{tabthird}0.27 &  \cellcolor{tabthird}15.96 &  \cellcolor{tabthird}0.59 &  \cellcolor{tabthird}0.28 & \cellcolor{tabsecond}18.51 &  \cellcolor{tabthird}0.65 &  \cellcolor{tabthird}0.26 &  \cellcolor{tabthird}12.44 &  \cellcolor{tabthird}0.39 &  \cellcolor{tabthird}0.54 &  \cellcolor{tabthird}16.36 &  \cellcolor{tabthird}0.46 &  \cellcolor{tabthird}0.30 &  \cellcolor{tabthird}15.99 &  \cellcolor{tabthird}0.62 &  \cellcolor{tabthird}0.34 \\
    NeuMan     & \cellcolor{tabsecond}18.42 & \cellcolor{tabsecond}0.58 & \cellcolor{tabsecond}0.20 & \cellcolor{tabsecond}18.39 & \cellcolor{tabsecond}0.64 & \cellcolor{tabsecond}0.19 &  \cellcolor{tabthird}17.66 & \cellcolor{tabsecond}0.66 & \cellcolor{tabsecond}0.24 & \cellcolor{tabsecond}19.05 & \cellcolor{tabsecond}0.66 & \cellcolor{tabsecond}0.21 &  \cellcolor{tabfirst}17.57 & \cellcolor{tabsecond}0.54 & \cellcolor{tabsecond}0.29 & \cellcolor{tabsecond}18.76 & \cellcolor{tabsecond}0.73 & \cellcolor{tabsecond}0.23 \\
    \midrule
    HUGS       &  \cellcolor{tabfirst}19.06 &  \cellcolor{tabfirst}0.67 &  \cellcolor{tabfirst}0.15 &  \cellcolor{tabfirst}19.16 &  \cellcolor{tabfirst}0.71 &  \cellcolor{tabfirst}0.16 &  \cellcolor{tabfirst}19.44 &  \cellcolor{tabfirst}0.73 &  \cellcolor{tabfirst}0.17 &  \cellcolor{tabfirst}19.48 &  \cellcolor{tabfirst}0.67 &  \cellcolor{tabfirst}0.18 & \cellcolor{tabsecond}17.45 &  \cellcolor{tabfirst}0.59 &  \cellcolor{tabfirst}0.27 &  \cellcolor{tabfirst}18.79 &  \cellcolor{tabfirst}0.76 &  \cellcolor{tabfirst}0.18
    
\\
    \bottomrule
    \end{tabular}  
    }
    \caption{Comparison of HUGS (ours) with previous work on the NeuMan dataset~\cite{jiang2022neuman} over \textbf{human-only} regions cropped using a tight bounding box. Performance is evaluated on PSNR, SSIM and LPIPS metrics.}
    \label{tab:neuman_human}
\end{table*}
\definecolor{tabfirst}{rgb}{1, 0.7, 0.7}
\definecolor{tabsecond}{rgb}{1, 0.85, 0.7}
\definecolor{tabthird}{rgb}{1, 1, 0.7}

\begin{table*}[htb!]
    \centering
    \resizebox{\textwidth}{!}{
    \begin{tabular}{c|ccc|ccc|ccc|ccc|ccc|ccc}
    \toprule
        & \multicolumn{3}{c|}{\textbf{377}} & \multicolumn{3}{c|}{\textbf{386}} & \multicolumn{3}{c|}{\textbf{387}} & \multicolumn{3}{c|}{\textbf{392}} & \multicolumn{3}{c|}{\textbf{393}} & \multicolumn{3}{c}{\textbf{394}}   \\
    \midrule
        & PSNR $\uparrow$ & SSIM $\uparrow$ & LPIPS $\downarrow$ & PSNR $\uparrow$ & SSIM $\uparrow$ & LPIPS $\downarrow$ & PSNR $\uparrow$ & SSIM $\uparrow$ & LPIPS $\downarrow$ & PSNR $\uparrow$ & SSIM $\uparrow$ & LPIPS $\downarrow$ & PSNR $\uparrow$ & SSIM $\uparrow$ & LPIPS $\downarrow$ & PSNR $\uparrow$ & SSIM $\uparrow$ & LPIPS $\downarrow$  \\
    \midrule

NeuralBody &                      29.11 & \cellcolor{tabsecond}0.97 & \cellcolor{tabsecond}0.04 &                      30.54 & \cellcolor{tabsecond}0.97 &  \cellcolor{tabthird}0.05 &                      27.00 &  \cellcolor{tabthird}0.95 &  \cellcolor{tabthird}0.06 &                      30.10 & \cellcolor{tabsecond}0.96 & \cellcolor{tabsecond}0.05 & \cellcolor{tabsecond}28.61 & \cellcolor{tabsecond}0.96 &  \cellcolor{tabthird}0.06 &  \cellcolor{tabthird}29.10 & \cellcolor{tabsecond}0.96 &  \cellcolor{tabthird}0.05 \\

HumanNerf  &  \cellcolor{tabthird}30.41 & \cellcolor{tabsecond}0.97 &  \cellcolor{tabfirst}0.02 & \cellcolor{tabsecond}33.20 &  \cellcolor{tabfirst}0.98 & \cellcolor{tabsecond}0.03 & \cellcolor{tabsecond}28.18 & \cellcolor{tabsecond}0.96 & \cellcolor{tabsecond}0.04 &  \cellcolor{tabthird}31.04 &  \cellcolor{tabfirst}0.97 &  \cellcolor{tabfirst}0.03 &                      28.31 & \cellcolor{tabsecond}0.96 & \cellcolor{tabsecond}0.04 & \cellcolor{tabsecond}30.31 & \cellcolor{tabsecond}0.96 &  \cellcolor{tabfirst}0.03  \\
MonoHuman  & \cellcolor{tabsecond}30.77 &  \cellcolor{tabfirst}0.98 &  \cellcolor{tabfirst}0.02 &  \cellcolor{tabthird}32.97 & \cellcolor{tabsecond}0.97 & \cellcolor{tabsecond}0.03 &  \cellcolor{tabthird}27.93 & \cellcolor{tabsecond}0.96 &  \cellcolor{tabfirst}0.03 & \cellcolor{tabsecond}31.24 &  \cellcolor{tabfirst}0.97 &  \cellcolor{tabfirst}0.03 &  \cellcolor{tabthird}28.46 & \cellcolor{tabsecond}0.96 &  \cellcolor{tabfirst}0.03 &                      28.94 & \cellcolor{tabsecond}0.96 & \cellcolor{tabsecond}0.04  \\
\midrule
\acronym   &  \cellcolor{tabfirst}30.80 &  \cellcolor{tabfirst}0.98 &  \cellcolor{tabfirst}0.02 &  \cellcolor{tabfirst}34.11 &  \cellcolor{tabfirst}0.98 &  \cellcolor{tabfirst}0.02 &  \cellcolor{tabfirst}29.29 &  \cellcolor{tabfirst}0.97 &  \cellcolor{tabfirst}0.03 &  \cellcolor{tabfirst}31.36 &  \cellcolor{tabfirst}0.97 &  \cellcolor{tabfirst}0.03 &  \cellcolor{tabfirst}29.80 &  \cellcolor{tabfirst}0.97 &  \cellcolor{tabfirst}0.03 &  \cellcolor{tabfirst}30.54 &  \cellcolor{tabfirst}0.97 &  \cellcolor{tabfirst}0.03  \\

    \bottomrule
    \end{tabular}  
    }
    \caption{Comparison of HUGS (ours) with the previous work on the ZJU Mocap dataset~\cite{peng2021neuralbody}. Performance is evaluated on PSNR, SSIM and LPIPS metric. HUGS achieves state-of-the-art performance across all scenes and all metrics.}
    \label{tab:zju}
\end{table*}

\paragraph{Disentanglement of the Human and the Scene.} HUGS allows for a disentangled represenation of the human and the scene by storing their Gaussian features separately. This allows us to move the human to different scenes. In \cref{fig:composition}, we show the composition of human captured in one scene into a different scene. We show additional video results in the supplemental material.

\subsection{Quantitative Results}
\label{sec:quantitative}
We compare the performance of our method with baselines such as NeRF-T~\cite{li2021neural}, HyperNeRF~\cite{park2021hypernerf} and the existing state-of-the-art methods -- NeuMan~\cite{jiang2022neuman} and Vid2Avatar~\cite{guo2023vid2avatar}.

In \cref{tab:neuman_human_scene}, we evaluate the reconstruction quality on the NeuMan dataset~\cite{jiang2022neuman} on three different metrics -- Peak Signal-to-Noise Ration (PSNR), SSIM~\cite{ssim} and LPIPS~\cite{zhang2018lpips}. NeRF-T and HyperNeRF are general dynamic scene reconstruction methods and do not specialize for humans. Therefore, they show poor reconstruction quality. On the other hand, NeuMan and Vid2Avatar employ specialized models for the human and the scene. NeuMan employs a NeRF-based~\cite{mildenhall2020nerf} approach for both scene and human modeling. Vid2Avatar utilizes an implicit SDF model and volume rendering for scene and human representation. Therefore, both NeuMan and Vid2Avatar show improved reconstruction quality. In comparison, our method achieves state-of-the-art performance across all the scenes and metrics except PSNR on the \textit{Bike} sequence where we show competitive performance.

In \cref{tab:neuman_human}, we further evaluate the reconstruction error but only on the regions containing the human. We first take a tight crop around the human region in the ground truth image. This crop is used over all the predictions, and the reconstruction error is evaluated over the cropped samples. It should be noted that we take rectangular crops of the region and do not use any segmentation mask since reconstruction metrics are highly sensitive to masks. Under this evaluation, we show state-of-the-art performance across all scenes and metrics except PSNR on the \textit{Jogging} sequence where we show competitive performance. 

In addition, we evaluate our method using the ZJU Mocap dataset~\cite{peng2021neuralbody} in ~\cref{tab:zju}. We compare with recent previous work that report their evaluation on this dataset which include NeuralBody~\cite{peng2021neuralbody}, HumanNerf~\cite{weng2022humannerf}, and MonoHuman~\cite{yu2023monohuman}. 



\label{sec:timing}
\begin{figure}[t]
    \centering
    \includegraphics[width=0.49\linewidth]{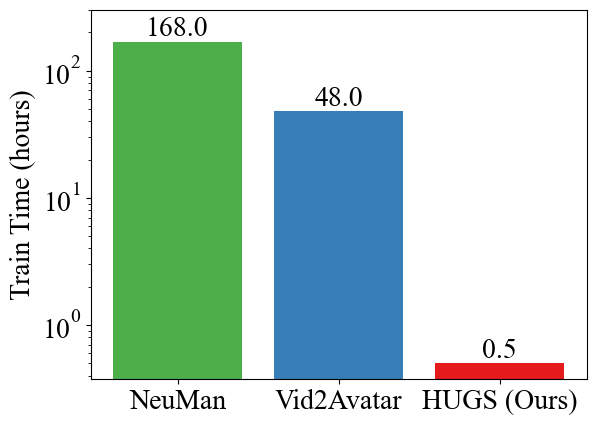}
    \includegraphics[width=0.49\linewidth]{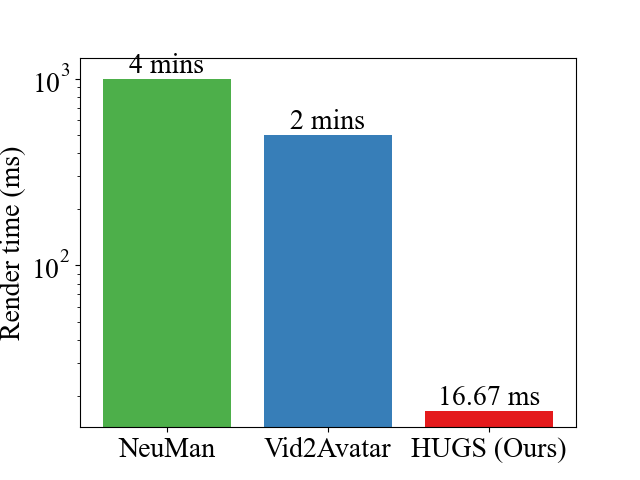}
    \caption{{Timing comparison for training in hours and rendering in milliseconds. y-axis is  on log-scale. HUGS outperforms previous methods by an order of magnitude. }} 
    \label{fig:timing}
\end{figure}
\begin{table*}[t]
    \centering
    \resizebox{\textwidth}{!}{
    \begin{tabular}{l|ccc|ccc|ccc|ccc|ccc|ccc}
    \toprule
        & \multicolumn{3}{c|}{\textbf{Seattle}} & \multicolumn{3}{c|}{\textbf{Citron}} & \multicolumn{3}{c|}{\textbf{Parking}} & \multicolumn{3}{c|}{\textbf{Bike}} & \multicolumn{3}{c|}{\textbf{Jogging}} & \multicolumn{3}{c}{\textbf{Lab}}   \\
    \midrule
        & PSNR $\uparrow$ & SSIM $\uparrow$ & LPIPS $\downarrow$ & PSNR $\uparrow$ & SSIM $\uparrow$ & LPIPS $\downarrow$ & PSNR $\uparrow$ & SSIM $\uparrow$ & LPIPS $\downarrow$ & PSNR $\uparrow$ & SSIM $\uparrow$ & LPIPS $\downarrow$ & PSNR $\uparrow$ & SSIM $\uparrow$ & LPIPS $\downarrow$ & PSNR $\uparrow$ & SSIM $\uparrow$ & LPIPS $\downarrow$  \\
    \midrule
    w/o LBS             &                      18.47 & \cellcolor{tabsecond}0.66 & \cellcolor{tabsecond}0.16 & \cellcolor{tabsecond}19.00 & \cellcolor{tabsecond}0.70 &  \cellcolor{tabfirst}0.16 & \cellcolor{tabsecond}19.13 & \cellcolor{tabsecond}0.72 & \cellcolor{tabsecond}0.19 & \cellcolor{tabsecond}19.73 &  \cellcolor{tabfirst}0.68 &  \cellcolor{tabfirst}0.18 &                      17.29 &  \cellcolor{tabthird}0.58 &  \cellcolor{tabthird}0.27 &  \cellcolor{tabthird}18.80 &  \cellcolor{tabfirst}0.76 &  \cellcolor{tabfirst}0.18 \\
    \midrule
    w/o Densify         & \cellcolor{tabsecond}18.91 &  \cellcolor{tabthird}0.65 & \cellcolor{tabsecond}0.16 &                      17.18 &  \cellcolor{tabthird}0.68 &  \cellcolor{tabthird}0.18 &  \cellcolor{tabthird}19.00 &  \cellcolor{tabthird}0.71 &  \cellcolor{tabthird}0.21 &  \cellcolor{tabfirst}19.92 & \cellcolor{tabsecond}0.67 & \cellcolor{tabsecond}0.19 & \cellcolor{tabsecond}17.63 &                      0.57 & \cellcolor{tabsecond}0.26 &  \cellcolor{tabfirst}18.98 &  \cellcolor{tabfirst}0.76 &  \cellcolor{tabfirst}0.18 \\
    \midrule
    w/o $\mathcal{L}^h$ &  \cellcolor{tabthird}18.87 &  \cellcolor{tabfirst}0.67 & \cellcolor{tabsecond}0.16 &  \cellcolor{tabthird}17.31 &                      0.67 &                      0.19 &                      17.76 &                      0.70 &                      0.23 &  \cellcolor{tabthird}19.63 &  \cellcolor{tabfirst}0.68 & \cellcolor{tabsecond}0.19 &  \cellcolor{tabfirst}18.23 &  \cellcolor{tabfirst}0.60 & \cellcolor{tabsecond}0.26 &                      18.75 &  \cellcolor{tabfirst}0.76 & \cellcolor{tabsecond}0.19 \\
    \midrule
    w/o Triplane        &                      18.70 &  \cellcolor{tabthird}0.65 &  \cellcolor{tabfirst}0.15 &                      17.12 &                      0.67 & \cellcolor{tabsecond}0.17 &                      18.78 &                      0.69 & \cellcolor{tabsecond}0.19 &                      19.52 &  \cellcolor{tabthird}0.66 &  \cellcolor{tabfirst}0.18 &  \cellcolor{tabthird}17.60 &                      0.56 &  \cellcolor{tabfirst}0.25 & \cellcolor{tabsecond}18.91 & \cellcolor{tabsecond}0.74 & \cellcolor{tabsecond}0.19 \\
    \midrule
    HUGS                &  \cellcolor{tabfirst}19.06 &  \cellcolor{tabfirst}0.67 &  \cellcolor{tabfirst}0.15 &  \cellcolor{tabfirst}19.16 &  \cellcolor{tabfirst}0.71 &  \cellcolor{tabfirst}0.16 &  \cellcolor{tabfirst}19.44 &  \cellcolor{tabfirst}0.73 &  \cellcolor{tabfirst}0.17 &                      19.48 & \cellcolor{tabsecond}0.67 &  \cellcolor{tabfirst}0.18 &                      17.45 & \cellcolor{tabsecond}0.59 &  \cellcolor{tabthird}0.27 &                      18.79 &  \cellcolor{tabfirst}0.76 &  \cellcolor{tabfirst}0.18
    \\ 
    \bottomrule
    \end{tabular}  
    }
    \caption{\textbf{Ablation study.} The performance is evaluated over human-only bounding box regions using PSNR, SSIM and LPIPS metrics. }
    \label{tab:neuman_ablation_human_only}
\end{table*}

\paragraph{Speed.} In ~\cref{fig:timing}, we compare the training and rendering time of our method with previous work. The use of 3DGS~\cite{kerbl3Dgaussians} speeds up our training and rendering times by a significant margin. We note that HUGS is {$96 \times$} faster than Vid2Avatar and {$336 \times $} faster than NeuMan  training within 30 minutes.
At rendering time, we do not rely on MLPs and only use the LBS weights, enabling higher frame rate.
Our method achieves {60 FPS} outperforming NeuMan by {${\sim}7600 \times$} and Vid2Avatar by {${\sim}3800 \times$}. We benchmark all the methods on a single GeForce 3090Ti GPU. \\


In summary, our model demonstrates efficiency in both training and rendering, delivering superior results compared to existing methods. 
Our model not only outperforms established NeRF and implicit-SDF based models but does so at orders of magnitude faster speeds.

\subsection{Ablation Experiments}
\begin{figure}[t]
    \centering
    \includegraphics[width=\linewidth]{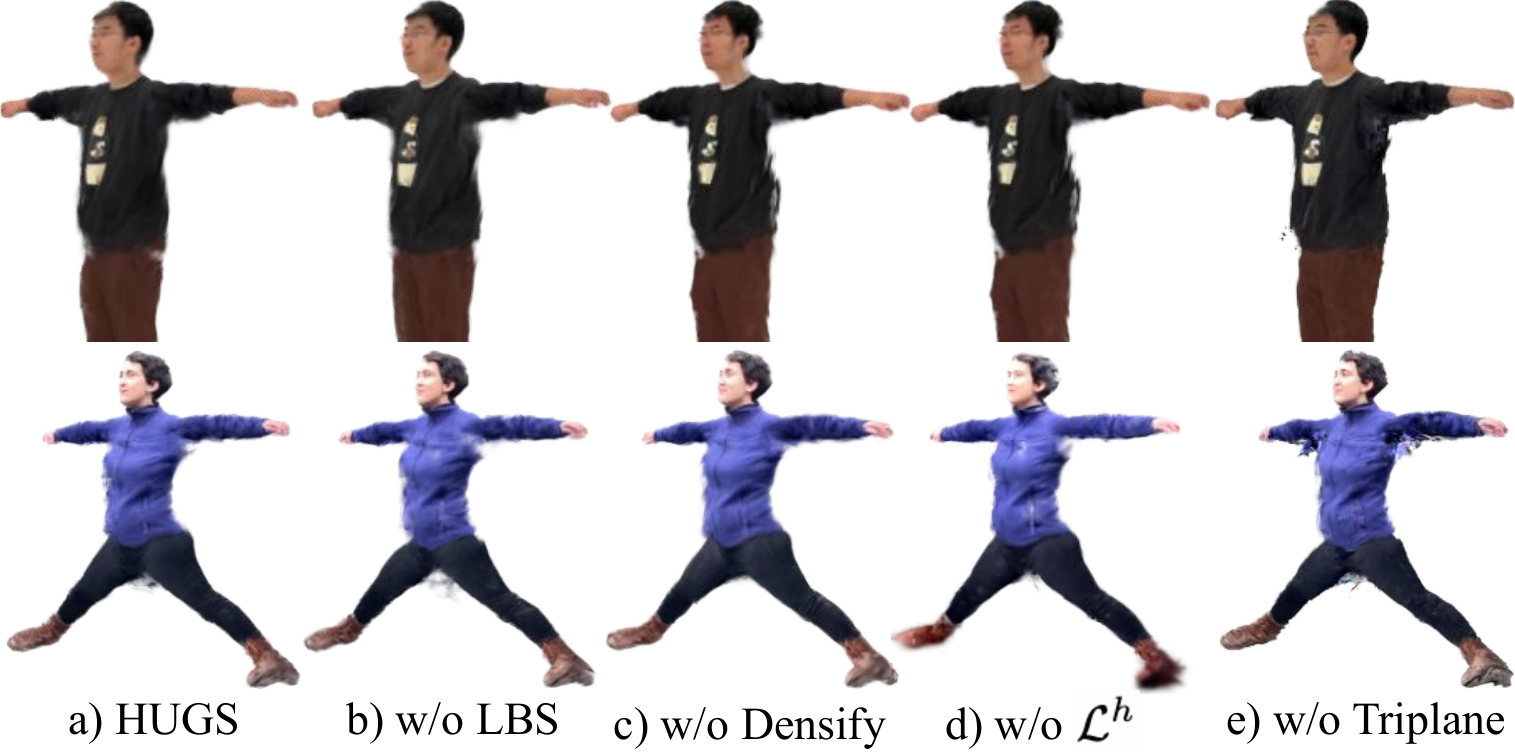}
    \caption{\textbf{Ablation study} showing the visualization of details captured in the human canonical shape under different ablations of our method.} 
    \label{fig:ablation}
\end{figure}

We show the effect of ablating over our method in ~\cref{fig:ablation}. We note that removing LBS from our full model results in floating artifacts that are mainly introduced in the corner region or the body. We also experiment by keeping the number of Human Gaussians to be fixed by disabling densification. This results in floaters around the edges (row 1, on the side of the shirt) since noisy Gaussians are not culled and large Gaussians are not split. 

Furthermore, we examine the effect of removing the loss on the human pixels $\mathcal{L}_h$. This results in loss of fine details in the human region as evident from the reconstruction of the shoes (row 2). In addition, removing the triplane+MLP and directly optimizing the 3DGS parameters results in noisy estimates. Please refer to supplemental material for a detailed ablation and analysis of our contributions.

In ~\cref{tab:neuman_ablation_human_only}, we show quantitative results on the NeuMan dataset by evaluating over only the human-regions by cropping it using a tight bounding box. We evaluate rendering quality using PSNR, SSIM and LPIPS metrics.










\section{Conclusion}
We have proposed \acronym, a new method for novel-view and novel-pose synthesis of a human embedded in the scene by bringing a deformable model into the Gaussian Splatting framework. The method is able to reconstruct human and scene representations from in-the-wild monocular videos containing a small number of frames (50-100). \acronym enables fast training (in 30 mins) and rendering (60 FPS), ~ $100\times$ faster than the previous methods~\cite{jiang2022neuman, guo2023vid2avatar}, while at the same time significantly improving rendering quality as measured by PSNR, SSIM and LPIPS metrics.

\paragraph{Limitations and Future Work.} HUGS is limited by the underlying shape model SMPL~\cite{SMPL:2015} and linear blend skinning that may not capture the general deformable structure of loose clothing such as dresses. In addition, HUGS is trained on in-the-wild videos that do not cover the pose-space of the human body. Future work will aim to alleviate these problems by modeling non-linear clothing deformation. In addition, the lack of data maybe alleviated by learning an appearance prior on human-poses using generative approaches such as GNARF~\cite{bergman2022gnarf} and AG3D~\cite{dong2023ag3d} or by distilling from image diffusion models~\cite{poole2022dreamfusion, lin2023magic3d}. Furthermore, our model does not account for environment lighting that may effect the composition of the human in a different scene with a different environment light which can be addressed by factoring out an illumination representation~\cite{verbin2022refnerf, ranjan2023facelit}. 




\small{
\noindent
{\bf Acknowledgements:} We thank Angelos Katharopoulos, Thomas Merth, Raviteja Vemulapalli, Barry Theobald, and Skyler Seto for their feedback on the manuscript and Wei Jiang for providing the details on NeuMan experiments.
}
{\small
\bibliographystyle{unsrt}
\bibliography{references}
}




\section{Appendix}

\subsection{Triplane-MLP Network Architecture}
\begin{figure*}[t]
    \centering
    \includegraphics[width=\linewidth]{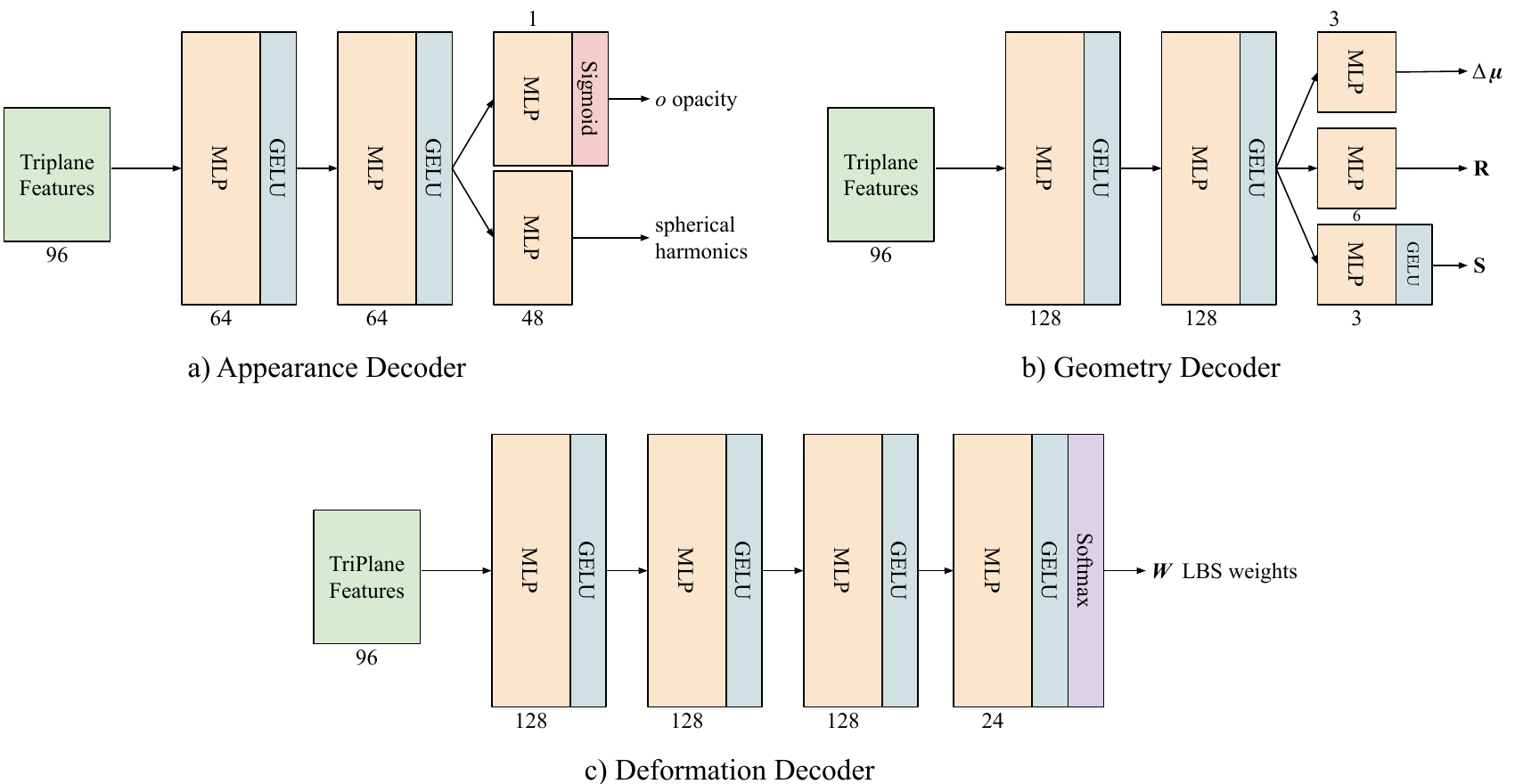}
    \vspace{-7mm}
    \caption{\textbf{\acronym model architecture.} Here we show the network architecture of decoder models. 
    Appearance decoder $D_A$ is a 2-layer MLP with GELU~\cite{hendrycks2016gelu} activations. It takes triplane features as input and outputs opacity \& spherical harmonics parameters. Sigmoid activation function is used for opacity to constrain the values between $[0, 1]$. Geometry decoder $D_G$ also uses 2-layer MLP with GELU activation. It outputs the $\Delta \bm{\mu}, \bm{R}, \bm{S}$ for each Gaussians. We use GELU activation to ensure $\bm{S} \ge 0$. We do not need a normalization activation for $R$ unlike Kerbl \etal~\cite{kerbl3Dgaussians} since we use 6D rotation representation. We use a 3-layer MLP for deformation decoder $D_D$. It outputs the LBS weights. We apply a low-temperature softmax activation function with $T=0.1$ to ensure $\sum_{k=1}^{n_k} W_{k,i} = 1$. The use of a low temperature assists in predicting unimodal distributions, as most Gaussians need to be assigned to a single bone transformation.
    }
    \label{fig:netarch}
    \vspace{-2ex}
\end{figure*}{}

We present a detailed overview of the triplane-MLP model in \cref{fig:netarch}, employed for representing human avatars. Our model utilizes lightweight MLP models to predict the appearance, geometry, and deformation parameters of individual Gaussians. This facilitates efficient model training within a short timeframe without significantly increasing the overall pipeline's computational burden. Through empirical investigation, we observed that the GELU activation function~\cite{hendrycks2016gelu} leads to quicker convergence. During the rendering phase, both triplane and MLP models are discarded once they predict the canonical human avatar and its deformation parameters. This design choice enables our method to achieve fast rendering at 60 FPS.

\paragraph{Appearance decoder ($D_A$)} comprises a 2-layer MLP with GELU activations~\cite{hendrycks2016gelu}. It takes triplane features as input and predicts opacity and spherical harmonics parameters. The opacity values are constrained within the range of $[0, 1]$ using a sigmoid activation function. Although no activation function is applied to the spherical harmonics parameters, the derived RGB values are clipped to be within $[0, 1]$ during the rasterization process.

\paragraph{Geometry decoder ($D_G$)} utilizes a 2-layer MLP with GELU activation, generating $\Delta \bm{\mu}, \bm{R}$, and $\bm{S}$ for each Gaussian. GELU activation ensures that $\bm{S} \ge 0$. Unlike Kerbl et al.~\cite{kerbl3Dgaussians}, we do not require a normalization activation for $\bm{R}$ as we employ a 6D rotation representation~\cite{Zhou_2019}.

\paragraph{Deformation decoder ($D_D$)} employs a 3-layer MLP to output LBS weights. We apply a low-temperature softmax activation function with $T=0.1$ to ensure $\sum_{k=0}^{n_k} \bm{W}_{k,i} = 1$. The use of a low temperature assists in predicting unimodal distributions, as most Gaussians need to be assigned to a single bone transformation.

\subsection{Loss implementation}
\paragraph{LBS regularization $\mathcal{L}_{\text{LBS}}$:} Given the limited set of training images, not regularizing LBS weights yields artifacts with unseen poses. Therefore, we apply regularization to ensure that the predicted LBS weights $\bm{W}$ closely align with those obtained from SMPL, employing an $\ell_2$ loss.

Specifically, to regularize the LBS weights $\bm{W}$, for individual Gaussians $\bm{p}_i$ we retrieve their $k=6$ nearest vertices on the \smpl mesh and take a distance-weighted average of their LBS weights to get $\hat{\bm{W}}$. The loss is $\mathcal{L}_{\text{LBS}} = \| \bm{W} - \hat{\bm{W}} \|_{\text{F}}^2$.

Following \cite{chen2021animatable,Zheng2020PaMIRPM} the distance-weighted average $\hat{\bm{W}_i}$ is obtained using:
\begin{align}
    \hat{\bm{W}_i} &= \sum_{j \in \mathcal{N}_i}  \frac{\omega_j}{\omega} \bm{W}_j \label{eq:lbsquery}\\
    \omega_j &= \exp \left( {-\frac{\| \bm{p}_i - \bm{p}_j \| \| \bm{W}_i - \bm{W}_j \|}{2\sigma^2}} \right) \\
    \omega &= \sum_{j \in \mathcal{N}_i} \omega_j
\end{align}

where $\mathcal{N}_i$ is the $k$ nearest neighbor of $\bm{p}_i$. We use the efficient CUDA $k$-nn implementation of PyTorch3D library~\cite{ravi2020pytorch3d}.


\subsection{Adaptive control of the number of 3D Gaussians}
We initialize the 3D Gaussians using a subdivided SMPL template with $n_v=110,210$ vertices. Despite the large number of vertices, the uniform subdivision across the body results in certain parts (e.g., face, hair, clothing) lacking sufficient points to represent high-frequency details. To address this issue, we adaptively control the number of 3D Gaussians during optimization, following a similar approach to Kerbl et al.~\cite{kerbl3Dgaussians}. The densification process starts after training the model for 3000 iterations, and subsequent densification steps are applied every 600 iterations.


\subsection{Ablation experiments}
\begin{figure*}[t]
    \centering
    \includegraphics[width=\linewidth]{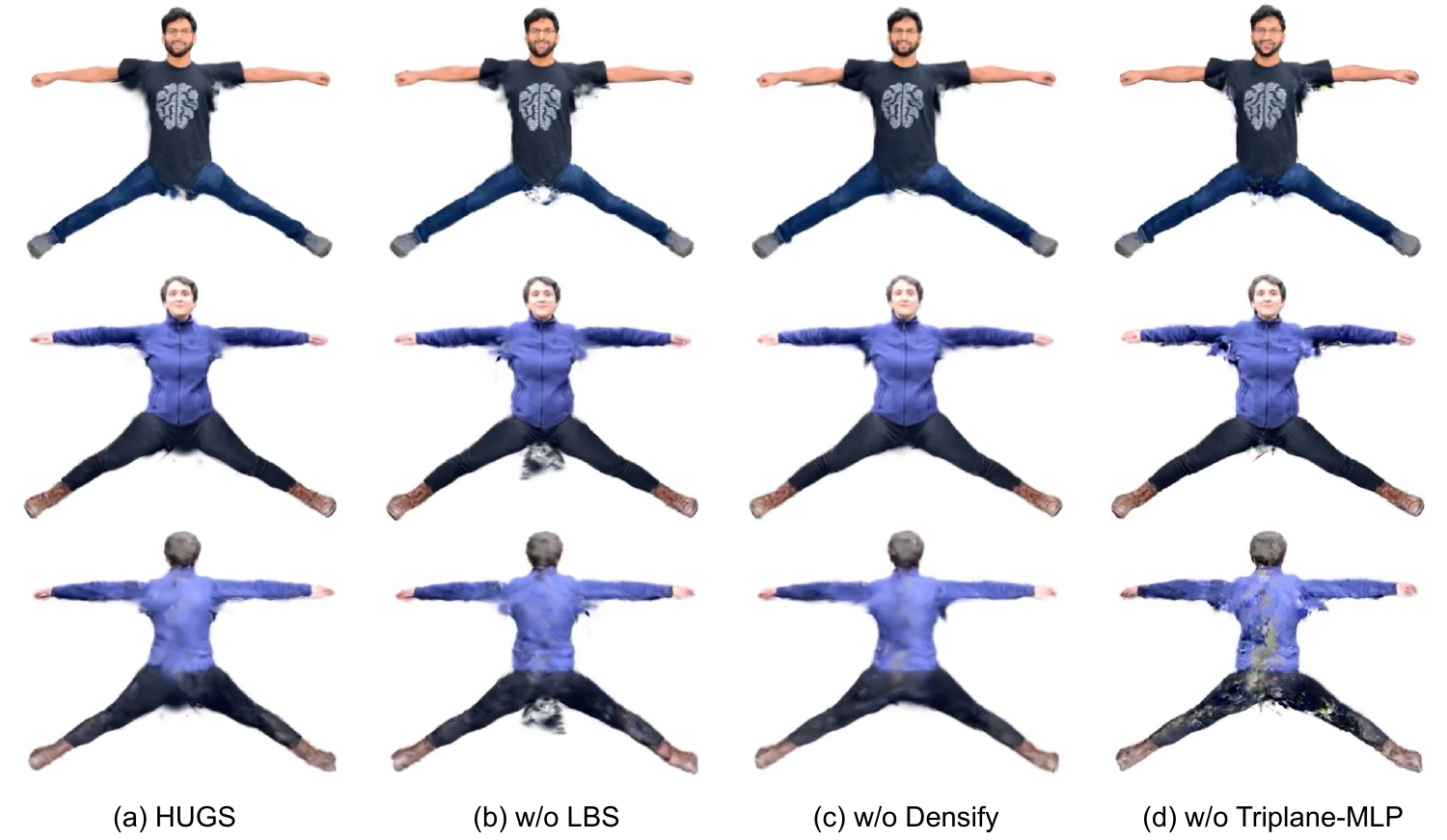}
    \caption{\textbf{Ablation study} showing the visualization of details captured in the human canonical shape under different ablations of our method.} 
    \label{fig:ablation_supmat}
\end{figure*}

In this section, we provide a detailed explanation of the ablation experiments.

\paragraph{w/o LBS baseline.} We replace the learned LBS weights with the SMPL LBS weights to assess the significance of learnable deformation. We obtain the SMPL LBS weights for a query point $\bm{p}_i$ using \cref{eq:lbsquery}. As the $k$-nn (k-nearest neighbors) approach can be ambiguous for points around the intersection of joints, utilizing SMPL weights may result in artifacts, particularly visible in areas around these intersections, as illustrated in (a) and (b) columns in \cref{fig:ablation_supmat}.

\paragraph{w/o Densify.} In this ablation experiment, we maintain a fixed number of Gaussians throughout the optimization process. The results of this experiment are depicted in the (a) and (c) columns in \cref{fig:ablation_supmat}. The \textit{w/o Densify} setting leads to certain Gaussians protruding from the body, resulting in a suboptimal reconstruction of details, as evidenced by the example of boot laces in the second row.

\paragraph{w/o triplane-MLP baseline.} In this ablation experiment, we directly optimize the 3D Gaussian parameters instead of learning them with a trilane-MLP model. To deform individual Gaussians, we utilize the SMPL LBS weights obtained using the query algorithm presented in \cref{eq:lbsquery}. The results of this experiment are displayed in the (a) and (d) columns in \cref{fig:ablation_supmat}. As each Gaussian is optimized independently, the per-Gaussian colors have a tendency to overfit to the training frames, leading to color artifacts. This results in poor quality on novel animation renderings and test frames. On the other hand, triplane-MLP provides implicit regularization to the color as a function of Gaussian position. Also, the learned appearance provides additional 3D supervision signal for the positions of the Gaussians.

\begin{figure*}[t]
    \centering
    \includegraphics[width=\linewidth]{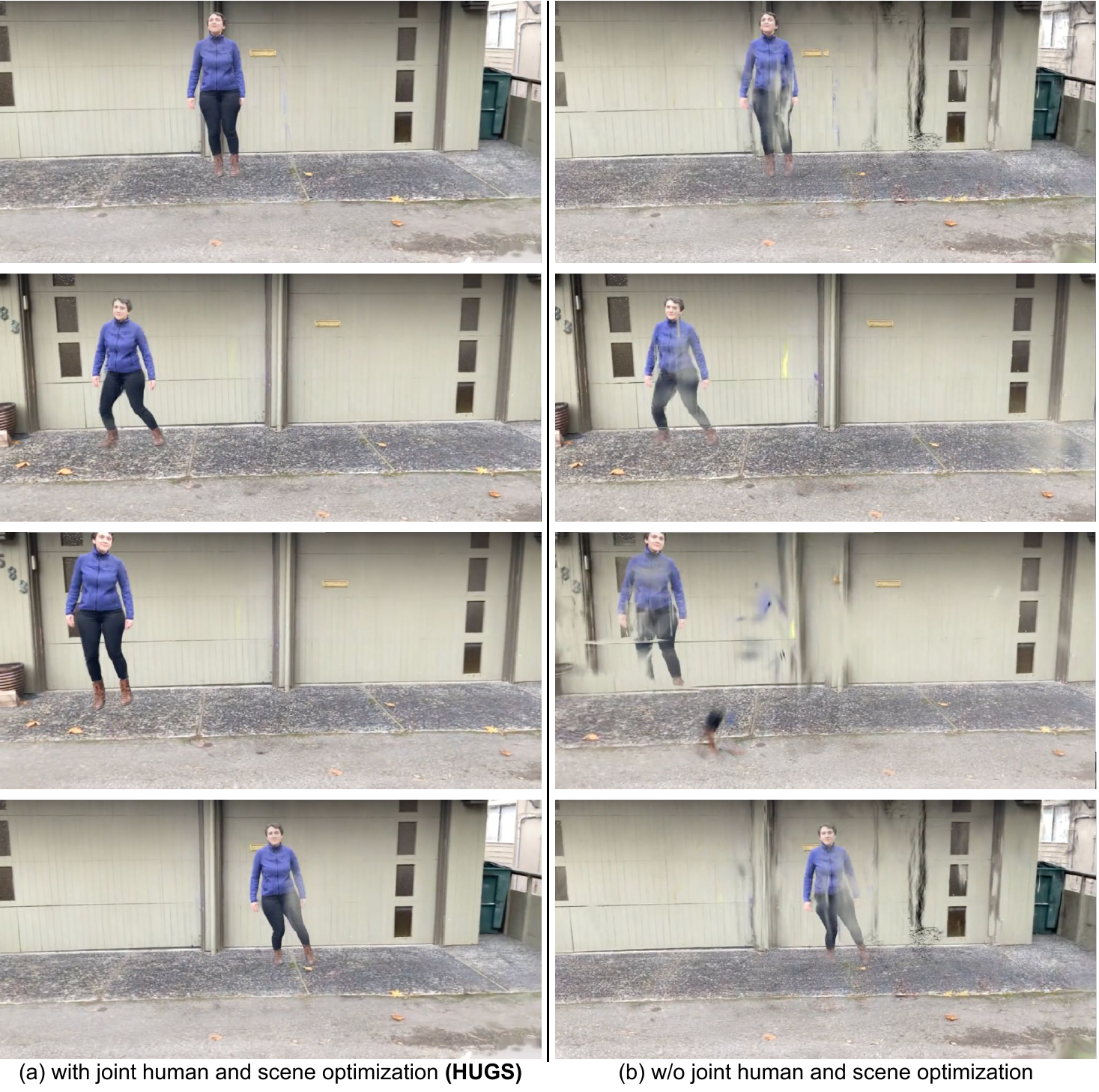}
    \caption{\textbf{Ablation study} showing the impact of jointly optimizing the human and scene models. Left column (a) shows the result of our method HUGS with joint optimization of human and scene Gaussians. Right column (b) shows optimizing the scene first by masking out the human regions and then optimizing the human Gaussians.} 
    \label{fig:ablation_joint_opt}
\end{figure*}

\paragraph{w/o Joint human \& scene optimization.} We assess the impact of jointly optimizing the human and scene models in this ablation experiment. Our final model, HUGS, represents human and scene Gaussians separately, however the optimization is performed jointly. An alternative approach involves initially optimizing the scene by masking out human regions and then optimizing the human Gaussians. This strategy aligns with the approach used in NeuMan~\cite{jiang2022neuman}. However, as illustrated in column (b) of \cref{fig:ablation_joint_opt}, this alternative strategy results in floating Gaussians in the scene due to sparse input views. On the contrary, when human and scene optimization is performed jointly, the human serves as a constraint for the scene reconstruction, mitigating the occurrence of floaters and yields cleaner rendered images as displayed in column (a) of \cref{fig:ablation_joint_opt}.


\subsection{Novel animation renderings}

\begin{figure*}[t]
    \centering
    \includegraphics[width=0.9\linewidth]{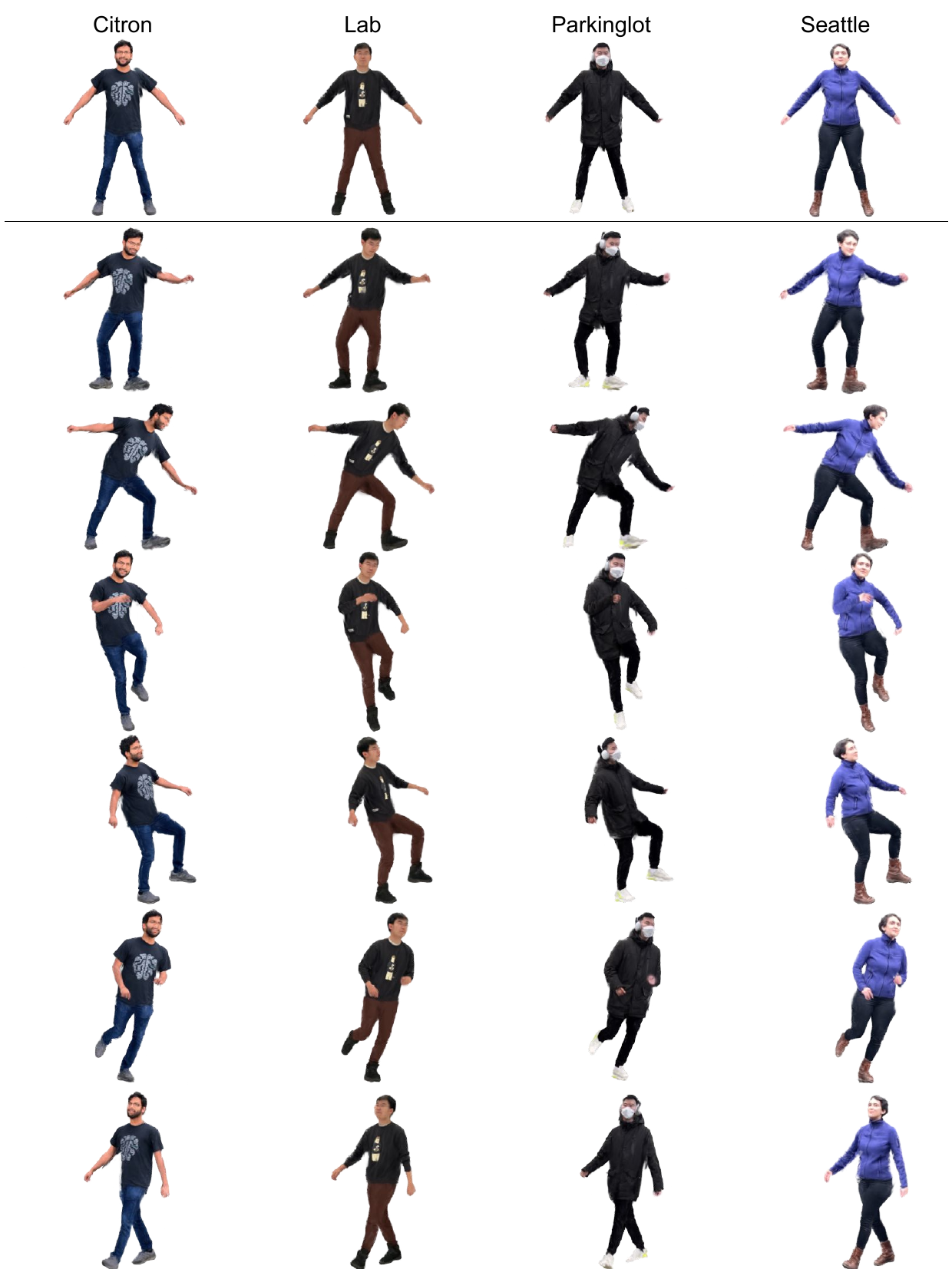}
    \caption{\textbf{Novel pose renderings} We demonstrate the novel pose renderings of subjects (top row) from the NeuMan dataset.} 
    \label{fig:posed_renders}
\end{figure*}

\begin{figure*}[t]
    \centering
    \includegraphics[width=\linewidth]{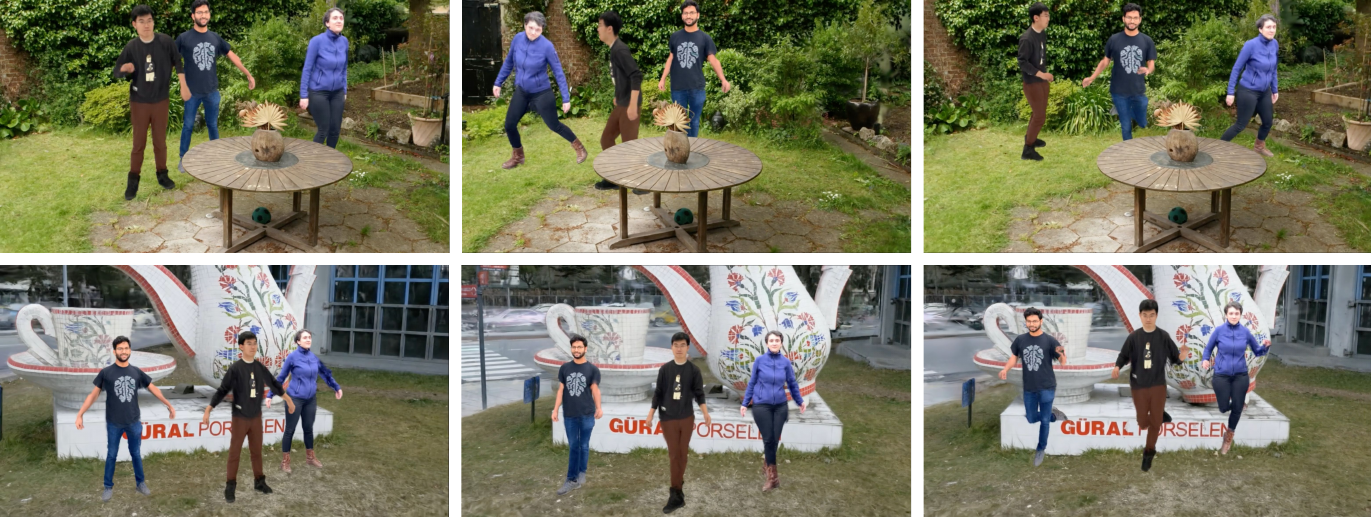}
    \caption{\textbf{Animation of multiple people in novel scenes.} Renderings obtained by transferring the Human Gaussians to different scenes.}
    \label{fig:novel_scene_pose}
\end{figure*}

In \cref{fig:posed_renders}, we present novel animation renderings featuring subjects from the NeuMan dataset~\cite{jiang2022neuman}. Additionally, \cref{fig:novel_scene_pose} showcases the composition of multiple animated subjects in various scenes, with poses obtained from the AMASS motion capture dataset~\cite{AMASS:ICCV:2019}. For a more extensive collection of video results demonstrating novel pose and scene animations, please refer to our supplementary webpage.

{}
{}

\end{document}